\documentclass[lettersize,journal]{IEEEtran}
\usepackage{amsmath,amsfonts}
\usepackage{algorithmic}
\usepackage{algorithm}
\usepackage{array}
\usepackage[caption=false,font=normalsize,labelfont=sf,textfont=sf]{subfig}
\usepackage{textcomp}
\usepackage{stfloats}
\usepackage{url}
\usepackage{verbatim}
\usepackage{graphicx}
\usepackage{cite}
\usepackage{multirow}
\usepackage{times}
\usepackage{latexsym}
\usepackage{soul}
\usepackage{url}
\usepackage{bm}
\usepackage[utf8]{inputenc}
\usepackage{graphicx}
\usepackage{mathrsfs}
\usepackage{amsthm}
\usepackage{booktabs}
\usepackage{amssymb}
\usepackage{color}
\usepackage{multicol}
\graphicspath{{Figure/}}
\theoremstyle{plain}
\newtheorem{theorem}{Theorem}

\theoremstyle{definition}

\theoremstyle{remark}

\begin{document}

\title{Steering the Noise: Turning Random Perturbations 
into Effective Descent for Memory-Efficient LLM Fine-Tuning}
\author{Feihu~Jin, Shipeng Cen, and~Ying~Tan,~\IEEEmembership{Senior Member,~IEEE}%
\thanks{The authors are with the School of Intelligence Science and Technology and the Institute for Artificial Intelligence, Peking University, Beijing 100190, China. Y. Tan is also with the State Key Laboratory of General Artificial Intelligence, Beijing 100190, China (e-mail: fhjin@stu.pku.edu.cn; censhipeng@pku.edu.cn, ytan@pku.edu.cn). Code: https://github.com/stan-anony/MeZO-GV}
}

\markboth{Journal of \LaTeX\ Class Files,~Vol.~14, No.~8, August~2021}%
{Shell \MakeLowercase{\textit{et al.}}: A Sample Article Using IEEEtran.cls for IEEE Journals}


\maketitle

\begin{abstract}
   Fine-tuning large language models (LLMs) achieves strong performance but is often limited by the memory overhead of backpropagation. Zeroth-order (ZO) optimization avoids this overhead by estimating gradients through forward passes alone, yet it typically converges slowly because random Gaussian perturbations yield high-variance gradient estimates in high-dimensional parameter spaces. In this paper, we propose a plug-and-play framework that turns random perturbations into more effective descent directions. The key idea is to draw a small pool of candidate perturbations, evaluate their loss values, and then select or combine those that are best aligned with the optimization objective. We develop two instantiations of this idea: MeZO-GV, which forms a guiding vector from the contrast between low-loss and high-loss perturbation groups, and MeZO-Greedy, which keeps the single best perturbation within a fixed evaluation budget. We theoretically show that both strategies yield a larger per-step reduction in the objective than standard ZO estimation, leading to improved convergence rates. Experiments on LLMs of different scales and architectures confirm that the proposed methods integrate naturally with existing ZO optimizers and consistently improve convergence speed and task accuracy. On OPT-13B, our approach outperforms all ZO baselines across 11 benchmarks and exceeds gradient-based methods on 9 of them, while retaining the memory efficiency of forward-only optimization.
\end{abstract}

\begin{IEEEkeywords}
Zeroth-order optimization, large language models, memory-efficient tuning, Black-box optimization.
\end{IEEEkeywords}

\section{Introduction}
\IEEEPARstart{T}{he} emergence of fine-tuning techniques for large language models (LLMs) has revolutionized natural language processing (NLP), enabling state-of-the-art performance in tasks such as text generation and question answering \cite{ref1,ref2}. However, as LLMs scale, the computational and memory demands of fine-tuning grow substantially. A primary bottleneck arises during backpropagation \cite{ref3}, which requires storing intermediate activations and gradients, leading to prohibitive memory overhead. While parameter-efficient fine-tuning (PEFT) methods \cite{ref4, ref5, ref6} reduce the number of trainable parameters, memory efficiency remains a critical constraint. For instance, experiments on OPT-13B \cite{ref7} indicate that full fine-tuning and PEFT still consume approximately $12\times$ and $6\times$ more GPU memory than inference, respectively \cite{ref8}.

\begin{figure*}[ht]
\centering
\begin{minipage}{0.3\textwidth}
\includegraphics[width=\linewidth]{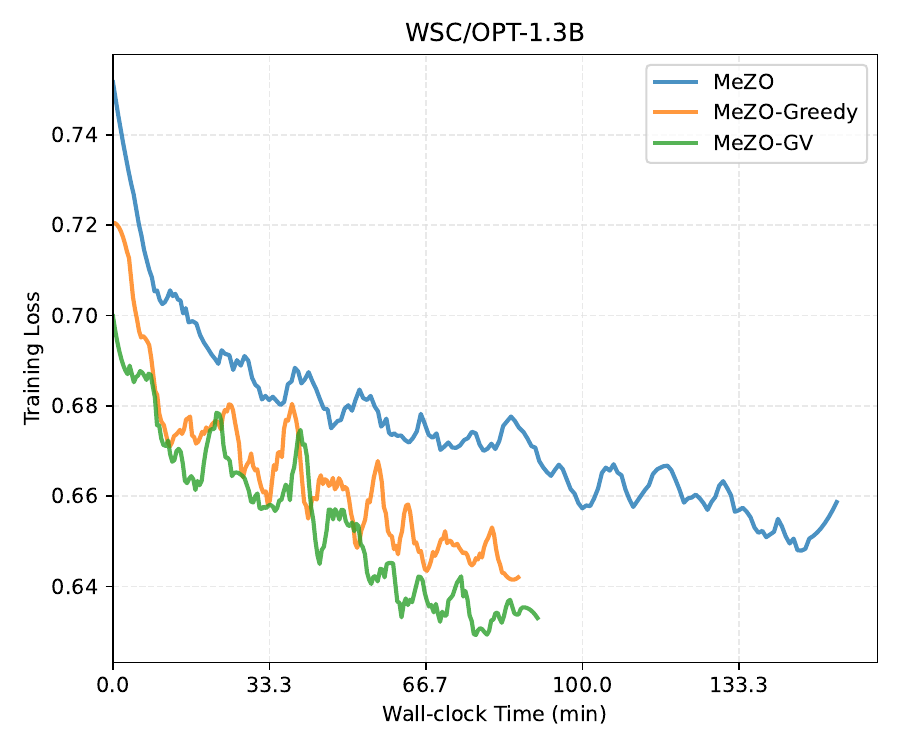}
\label{fig:wsc_loss}
\end{minipage}
\hfill
\begin{minipage}{0.3\textwidth}
\includegraphics[width=\linewidth]{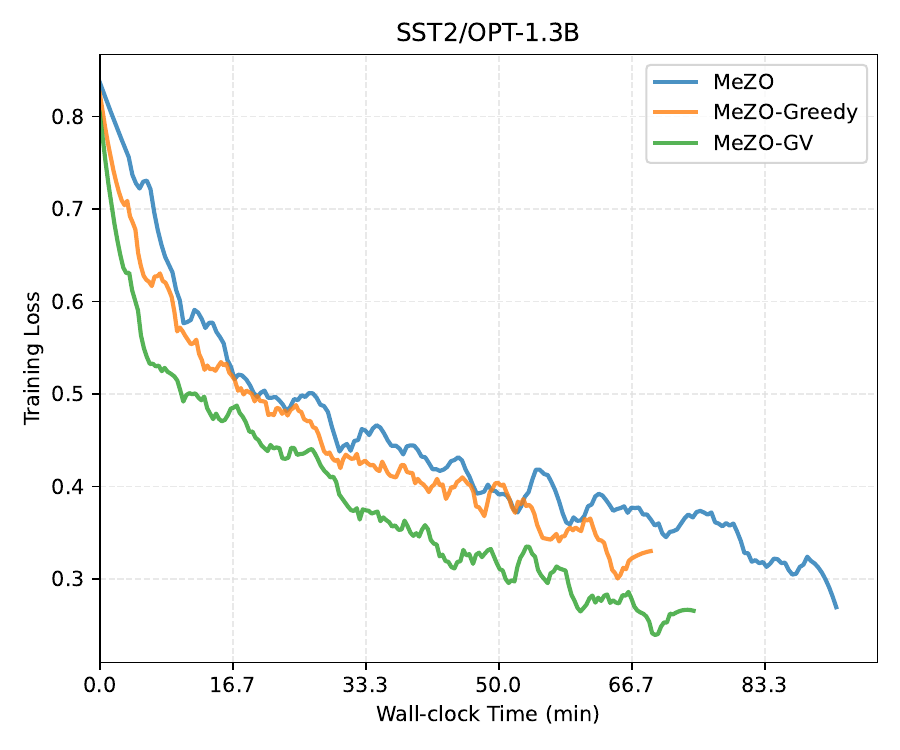}
\label{fig:sst2_ft_loss}
\end{minipage}
\hfill
\begin{minipage}{0.3\textwidth}
\includegraphics[width=\linewidth]{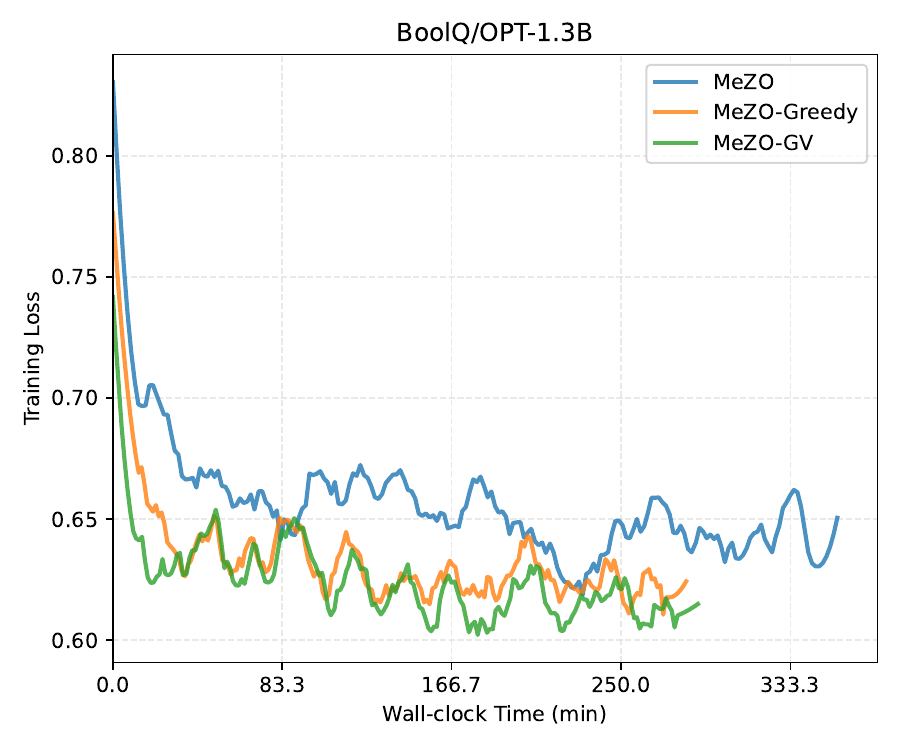}
\label{fig:boolq_ft}
\end{minipage}
\caption{Training loss curves for WSC, SST-2, and BoolQ on OPT-1.3B, plotted against wall-clock time. MeZO-Greedy and MeZO-GV integrate with MeZO and converge faster under the same evaluation budget. All experiments use a batch size of 16.}
\label{opt_13_loss}
\end{figure*}

Zeroth-order (ZO) optimization has emerged as a practical alternative that replaces backpropagation with gradient estimation through forward passes alone \cite{ref8}. Because ZO methods do not store intermediate activations, they greatly reduce memory usage. Recent work has improved ZO convergence through sparse perturbation strategies \cite{ref9}, hybrid frameworks combining ZO with adaptive optimizers \cite{ref10} or Hessian-aware estimation \cite{ref11}, and integration with PEFT \cite{ref12} to further reduce the number of trainable parameters.

Despite these advances, a fundamental bottleneck remains. The $n$-SPSA framework \cite{ref13} estimates the gradient by sampling $n$ independent random perturbations, computing a separate SPSA gradient estimate for each, and averaging the results. MeZO \cite{ref8} uses a single random perturbation per step and requires only two forward passes to produce one SPSA gradient estimate. Because each perturbation is drawn from an isotropic Gaussian distribution, its expected alignment with the true gradient scales as $O(1/d)$, where $d$ is the number of parameters. For billion-parameter LLMs, this means each random perturbation captures only a vanishingly small fraction of the gradient information, resulting in high-variance updates and slow convergence.

A natural idea to mitigate this issue is to increase $n$, i.e., draw multiple perturbations per step and average their SPSA estimates. When $n > 1$, $n$-SPSA reduces the variance of the gradient estimator by a factor of $n$ compared to MeZO ($n=1$), but each individual estimate still relies on a random direction with $O(1/d)$ alignment. In other words, increasing $n$ in $n$-SPSA yields a more stable estimate of the same low-quality direction, without improving the direction itself. The additional $2(n-1)$ forward passes are spent on variance reduction rather than on finding a better descent direction. Our key insight is that the loss values obtained during these extra forward evaluations carry rich information about the local loss landscape, and this information can be exploited to actively construct a better descent direction rather than merely averaging noisy estimates.

We present a plug-and-play zeroth-order optimization framework that turns random perturbations into more effective descent directions by leveraging loss feedback from multiple candidates. Given the same budget of $2n$ forward passes per step, $n$-SPSA independently computes $n$ SPSA estimates and averages them, whereas our framework evaluates $n$ candidate perturbations and uses their loss values to identify which directions are most aligned with the optimization objective. This transforms the role of additional forward passes from passive variance reduction to active directional selection. As shown in Figure~\ref{opt_13_loss}, under the same total number of forward evaluations, our methods converge noticeably faster than both MeZO, reducing end-to-end training time.

We develop two complementary strategies within this framework. MeZO-GV constructs a guiding vector by contrasting the centroid of low-loss perturbations with that of high-loss perturbations. This contrast amplifies the gradient signal: the low-loss perturbations tend to point along the descent direction while the high-loss ones point toward ascent, so their difference doubles the effective gradient projection. MeZO-Greedy takes a simpler selection-based approach, keeping only the single perturbation that produces the lowest loss from a candidate pool. We theoretically show that MeZO-GV achieves an $O(M)$ per-iteration speedup over standard MeZO, while MeZO-Greedy achieves an $O(\log M)$ speedup, where $M$ is the total number of forward passes per step. Both strategies are plug-and-play modules that integrate naturally with existing ZO optimizers such as MeZO \cite{ref8} and SubZero \cite{ref30}, without increasing memory consumption.

Extensive experiments across models ranging from OPT-1.3B to OPT-30B and Llama2-7B to Llama2-13B validate the effectiveness of our approach. On OPT-13B, GV-based methods achieve the best zeroth-order performance on all 11 benchmarks and outperform gradient-based fine-tuning on 9 of 11 tasks, while retaining the memory efficiency of forward-only optimization. Across all model scales and tuning schemes, our methods consistently outperform $n$-SPSA under the same computational budget, confirming that loss-guided directional selection is fundamentally more effective than variance reduction through averaging. 
\section{Related Work}
\subsection{Memory-Efficient ZO-SGD (MeZO)}
The Simultaneous Perturbation Stochastic Approximation (SPSA) \cite{ref13} is a zeroth-order method that approximates the gradient of a scalar-valued function \( f(\bm{x}) \), \( \bm{x} \in \mathbb{R}^d \), using finite differences along random Gaussian directions:
\begin{equation}
    \hat{\nabla} f(\bm{x}) = \frac{1}{q} \sum_{i=1}^q \frac{f(\bm{x} + \mu \bm{u}_i) - f(\bm{x} - \mu \bm{u}_i)}{2\mu} \bm{u}_i,
    \label{eq:rge}
\end{equation}
where \( q \) is the number of function evaluations, \( \mu > 0 \) is the perturbation step size, and \( \bm{u}_i \sim \mathcal{N}(\bm{0}, \bm{I}) \) are random direction vectors. As \( \mu \to 0 \), the finite difference approaches the directional derivative \( f'(\bm{x}, \bm{u}) = \bm{u}^\top \nabla f(\bm{x}) \), giving an unbiased gradient estimator:
\begin{equation}
    \mathbb{E}_{\bm{u}}[f'(\bm{x}, \bm{u}) \bm{u}] = \mathbb{E}_{\bm{u}}[\bm{u}\bm{u}^\top \nabla f(\bm{x})] = \nabla f(\bm{x}),
    \label{eq:unbiased}
\end{equation}
which makes SPSA well suited for high-dimensional optimization, including LLM fine-tuning.

Given a labeled dataset \( \mathcal{D} = \{(\bm{x}_i, y_i)\}_{i=1}^{|\mathcal{D}|} \), a minibatch \( \mathcal{B} \subset \mathcal{D} \), and a loss function \( \mathcal{L}(\bm{\theta}; \mathcal{B}) \) with parameters \( \bm{\theta} \in \mathbb{R}^d \), the SPSA gradient estimate takes the form:
\begin{equation}
    \hat{\nabla} \mathcal{L}(\bm{\theta}; \mathcal{B}) = \frac{\mathcal{L}(\bm{\theta} + \epsilon \bm{z}; \mathcal{B}) - \mathcal{L}(\bm{\theta} - \epsilon \bm{z}; \mathcal{B})}{2\epsilon} \bm{z},
    \label{eq:spsa}
\end{equation}
where \( \bm{z} \sim \mathcal{N}(\bm{0}, \bm{I}) \) is a random perturbation vector and \( \epsilon > 0 \) is the perturbation scale.
Since \( \hat{\nabla} \mathcal{L}(\bm{\theta}; \mathcal{B}) \approx \bm{z}\bm{z}^\top \nabla \mathcal{L}(\bm{\theta}; \mathcal{B}) \) requires only two forward passes, it enables memory-efficient optimization and forms the basis of Zeroth-Order Stochastic Gradient Descent (ZO-SGD):
\begin{equation}
    \bm{\theta}_{t+1} = \bm{\theta}_t - \eta \hat{\nabla} \mathcal{L}(\bm{\theta}; \mathcal{B}_t),
    \label{eq:zo_sgd}
\end{equation}
where \( \mathcal{B}_t \) is the \( t \)-th minibatch and \( \eta \) is the learning rate. ZO-SGD reduces the memory overhead of backpropagation by replacing exact gradients with SPSA estimates.

\subsection{Gradient-free Optimization of LLMs}
Recent gradient-free methods have used CMA-ES \cite{ref14} to optimize continuous prompt vectors in black-box tuning, though they often suffer from instability and slow convergence \cite{ref15,ref16}. To address these issues, \cite{ref17} introduced a gradient-free framework for low-rank adaptation.

Zeroth-order (ZO) optimization is another widely used gradient-free approach, particularly when gradient computation is unavailable \cite{ref18,ref19,ref20}. It has been applied to distributed optimization \cite{ref21}, adversarial example generation \cite{ref22, ref23}, and optimization without explicit gradient estimation \cite{ref24, ref25, ref26}. Recent ZO methods for LLM fine-tuning have demonstrated significant reductions in GPU memory usage \cite{ref8, ref27, ref28}.
Notable directions include variance reduction through larger batch sizes \cite{ref27, ref29}, sparse perturbation strategies \cite{ref9, ref10}, Hessian-aware estimation \cite{ref11}, and parameter-efficient approaches based on subspace mapping and PEFT \cite{ref4, ref6, ref12}.
\section{Our Proposed Method}
A key limitation of standard ZO methods is that random perturbation directions are poorly aligned with the true gradient. When $\bm{z} \sim \mathcal{N}(\bm{0}, \bm{I})$ is sampled in $\mathbb{R}^d$, its expected alignment with the gradient scales as $O(1/d)$, which becomes very small for billion-parameter LLMs , as analyzed in Section~\ref{theory}.

We propose to turn random perturbations into more effective descent directions by leveraging loss evaluations from multiple candidates. We introduce two complementary strategies: MeZO-GV aggregates information from multiple perturbations to form a guiding vector, yielding an $O(M)$ per-iteration speedup; MeZO-Greedy selects the perturbation with the lowest loss, yielding an $O(\log M)$ speedup. Both strategies are plug-and-play modules that can be combined with existing ZO frameworks. We integrate them into MeZO \cite{ref8} and SubZero \cite{ref30}.
\subsection{Memory-efficient ZO with Guiding Vector}
We propose MeZO-GV, which uses loss evaluations from multiple perturbations to construct a guiding vector that achieves steeper descent than a single random perturbation.

Consider $M-2$ independent random perturbations $\{\bm{z}^{(i)}\}_{i=1}^{M-2}$ sampled from $\mathcal{N}(\bm{0}, \bm{I})$. Under a first-order Taylor expansion, the perturbed loss at $\theta + \epsilon \bm{z}^{(i)}$ approximates:
\begin{equation}
    \mathcal{L}(\theta + \epsilon \bm{z}^{(i)}) \approx \mathcal{L}(\theta) + \epsilon \nabla\mathcal{L}(\theta)^\top \bm{z}^{(i)}.
\end{equation}
Perturbations that produce low losses therefore have a large projection onto the steepest descent direction, while those producing high losses point toward ascent. By contrasting these two groups, we can construct a direction that consistently points from high-loss regions toward low-loss regions.

Let $K = \lfloor \alpha (M-2) \rfloor$, where $\alpha \in (0, 0.5]$ is the selection ratio. After evaluating all $M-2$ perturbations, we split them into two sets by loss value: $S_{\text{low}}$, the $K$ perturbations with the lowest losses, and $S_{\text{high}}$, the $K$ perturbations with the highest losses. The guiding vector is the difference between their centroids:
\begin{equation}
\bm{v} = \underbrace{\frac{1}{K} \!\sum_{\bm{z} \in S_{\text{low}}}\! \bm{z}}_{\text{low-loss mean}} - \underbrace{\frac{1}{K} \!\sum_{\bm{z} \in S_{\text{high}}}\! \bm{z}}_{\text{high-loss mean}}.
\label{eq:guiding_vector}
\end{equation}
This construction orients $\bm{v}$ toward the direction of steepest local loss reduction. As shown in Section~\ref{theory}, it yields a per-iteration speedup of $O(M)$ over standard MeZO.

Using the guiding vector $\bm{v}$, we estimate the descent step via finite differences along this informed direction:
\begin{equation}
\hat{\nabla} \mathcal{L}(\bm{\theta}; \mathcal{B}) = \frac{\mathcal{L}(\bm{\theta} + \epsilon \bm{v}; \mathcal{B}) - \mathcal{L}(\bm{\theta} - \epsilon \bm{v}; \mathcal{B})}{2\epsilon} \bm{v}.
\end{equation}
Because $\bm{v}$ is constructed to have a larger directional derivative than a random vector, this estimator captures more of the loss landscape information per step.

\subsection{Memory-efficient ZO with Greedy Perturbation}
As a complementary strategy, MeZO-Greedy takes a selection-based approach: it identifies the single most promising perturbation from a candidate pool to maximize the immediate loss reduction.

Given $M-2$ candidate perturbations $\{\bm{z}^{(i)}\}_{i=1}^{M-2}$ sampled from $\mathcal{N}(\bm{0}, \bm{I})$, the greedy strategy selects the perturbation $\bm{z}^*$ that minimizes the perturbed loss:
\begin{equation}
\bm{z}^* = \arg\min_{1 \leq i \leq M-2} \mathcal{L}(\bm{\theta} + \epsilon \bm{z}^{(i)}; \mathcal{B}).
\end{equation}
Under a first-order Taylor approximation, this amounts to choosing the candidate with the largest projection along the steepest descent direction. As shown in Section~\ref{theory}, this yields a per-iteration speedup of $O(\log M)$ over standard MeZO.

Once $\bm{z}^*$ is identified, the gradient is estimated via symmetric finite differences:
\begin{equation}
\hat{\nabla} \mathcal{L}(\bm{\theta}; \mathcal{B}) = \frac{\mathcal{L}(\bm{\theta} + \epsilon \bm{z}^*; \mathcal{B}) - \mathcal{L}(\bm{\theta} - \epsilon \bm{z}^*; \mathcal{B})}{2\epsilon} \bm{z}^*.
\end{equation}
The parameters are then updated according to Equation \ref{eq:zo_sgd}.

\section{Theoretical Analysis}\label{theory}
  
In this section, we analyze the convergence properties and computational efficiency of MeZO-Greedy and MeZO-GV. Our analysis builds on the Greedy Descent Framework \cite{ref46}, which establishes a quantitative link between the quality of the perturbation direction and the convergence rate of zeroth-order optimization.

\subsection{Convergence via Descent Efficiency}

We consider unconstrained optimization of a convex and $L$-smooth loss function $\mathcal{L}(\theta)$ over $\mathbb{R}^d$. At each iteration $t$, the parameters are updated as $\theta_{t+1} = \theta_t - \eta_t \hat{g}_t$, where $\hat{g}_t$ is derived from a unit guiding vector $v_t$ ($\|v_t\|=1$). The quality of this update is measured by the effective descent coefficient $C_t$, defined as the squared cosine similarity between $v_t$ and the gradient $g_t = \nabla \mathcal{L}(\theta_t)$:
\begin{equation}
C_t := \cos^2(g_t, v_t) = \frac{(g_t^\top v_t)^2}{\|g_t\|^2 \|v_t\|^2}.
\end{equation}

\begin{theorem}[Convergence Rate]
\label{thm:convergence_rate}
Let $R = \sup_{\theta \in \mathcal{S}_0} \|\theta - \theta^*\|$ be the diameter of the initial level set $\mathcal{S}_0 = \{\theta : \mathcal{L}(\theta) \leq \mathcal{L}(\theta_0)\}$. For an $L$-smooth convex function, the expected suboptimality gap after $T$ iterations is bounded by:
\begin{equation}
    \mathbb{E}[\mathcal{L}(\theta_T) - \mathcal{L}(\theta^*)] \leq \frac{2 L R^2}{\sum_{t=0}^{T-1} \mathbb{E}[C_t]}.
\end{equation}
\end{theorem}

The convergence rate is thus inversely proportional to the cumulative expected descent coefficient $\sum \mathbb{E}[C_t]$. Any strategy that increases $\mathbb{E}[C_t]$ therefore produces a faster reduction in the objective per iteration.

\subsection{Analysis of Descent Efficiency}

We now analyze $\mathbb{E}[C_t]$ in the high-dimensional regime $d \gg 1$. Without loss of generality, we align the coordinate system so that the gradient $g = \nabla\mathcal{L}(\theta_t)$ points along the first basis vector, i.e., $g = \|g\| e_1$.

\paragraph{Standard MeZO.}
When $z \sim \mathcal{N}(0, I_d)$ is sampled at random, the descent coefficient is $C_t = z_1^2 / \sum_{j=1}^d z_j^2$. By the exchangeability of Gaussian components, its expectation is:
\begin{equation}
    \mathbb{E}[C_t^{\text{MeZO}}] = \frac{1}{d}.
\end{equation}
This $d^{-1}$ scaling explains the slow convergence of standard ZO methods on high-dimensional LLMs.

\paragraph{MeZO-Greedy.}
The greedy strategy samples $M-2$ candidates and keeps the one with the lowest loss. This corresponds to selecting the minimum projection along the first dimension, $Y_{(1)} = \min_{i} z_1^{(i)}$. By extreme value theory, $\mathbb{E}[Y_{(1)}^2] \approx 2\log (M-2)$. Since the orthogonal components remain unbiased, the expected descent coefficient is:
\begin{equation}
    \mathbb{E}[C_t^{\text{Greedy}}] \approx \frac{2\log (M-2)}{d}.
\end{equation}
This gives a logarithmic speedup of $\mathcal{O}(\log M)$ over standard MeZO. Although derived asymptotically, this trend holds empirically even for small $M$.

\paragraph{MeZO-GV.}
MeZO-GV takes the bottom $K = \lfloor \alpha (M-2) \rfloor$ perturbations (lower loss) and subtracts the top $K$ perturbations (higher loss). The resulting guiding vector has a signal component $\mathbb{E}[(v_{\text{raw}})_1^2] \approx 4(\mu_\alpha^-)^2$, where $\mu_\alpha^- = -\phi(\Phi^{-1}(\alpha))/\alpha$ is the truncated mean, and a noise energy of $2d/K$ from averaging. This gives:
\begin{equation}
    \mathbb{E}[C_t^{\text{GV}}] \approx \frac{2K(\mu_\alpha^-)^2}{d}.
\end{equation}
The speedup factor is $2K(\mu_\alpha^-)^2$ over standard MeZO, scaling as $\mathcal{O}(M)$ since $K = \lfloor\alpha(M-2)\rfloor$.

Table \ref{zo_analysis} summarizes the theoretical properties. The per-iteration convergence ordering is $\mathcal{O}(M) > \mathcal{O}(\log M) > \mathcal{O}(1)$, confirming the advantage of the guiding vector approach. We also report the compute-normalized speedup, defined as the per-iteration speedup divided by the forward-pass overhead $M/2$. For our default setting with $M=4$, $\alpha=0.5$, $K=1$, and $\mu_{0.5}^- \approx -0.798$, this ratio is approximately $0.64$, indicating slightly lower per-forward-pass efficiency than MeZO. However, the improved directional alignment reduces optimization variance and leads to earlier convergence, as confirmed by wall-clock experiments in Section~\ref{sec:wallclock}, where MeZO-GV achieves up to 29\% faster training.

\begin{table}[h]
\caption{Comparison of Expected Descent Efficiency and Speedup Factors}
\centering
\scriptsize
\begin{tabular}{lccc}
\hline
\textbf{Method} & $\mathbb{E}[C_t]$ & \textbf{Speedup} & \textbf{FP/Step} \\ \hline
Standard MeZO & $\frac{1}{d}$ & $1\times$ & $2$ \\
MeZO-Greedy & $\frac{2\log (M\!-\!2)}{d}$ & $2\log (M\!-\!2)$ & $M$  \\
MeZO-GV & $\frac{2K(\mu_\alpha^-)^2}{d}$ & $2K(\mu_\alpha^-)^2$ & $M$  \\ \hline
\end{tabular}
\label{zo_analysis}
\end{table}

\begin{table}[h]
\centering
\tiny
\caption{Hyperparameters for OPT and Llama2 (Batch Size: 16, Subspace Freq.: \{500, 1000, 2000\})}
\label{tab:unified_params}
\begin{tabular}{llcccc}
\hline
\textbf{Tuning} & \textbf{Algorithm} & \textbf{LR} & $\epsilon$ & $k$ & \textbf{Rank} \\ \hline
\multirow{3}{*}{\textbf{FT}} & MeZO / SubZero & \{1e-7, 2e-7, 5e-7\} & 1e-3 & -- & \{32, 64\} \\
 & MeZO-GV / SubZero-GV & \{1e-7, 2e-7, 3e-7, 5e-7\} & 1e-3 & 4 & \{32, 64\} \\
 & SGD & \{1e-4, 1e-3, 5e-3\} & -- & -- & -- \\ \hline
\multirow{2}{*}{\textbf{LoRA}} & MeZO / SubZero & \{3e-5, 5e-5, 1e-4\} & 1e-2 & -- & \{32, 64\} \\
 & MeZO-GV / SubZero-GV & \{3e-5, 5e-5, 1e-4\} & 1e-2 & 4 & \{32, 64\} \\ \hline
\multirow{2}{*}{\textbf{Prefix}} & MeZO / SubZero & \{1e-3, 5e-3, 1e-2\} & 1e-1 & -- & \{8, 16\} \\
 & MeZO-GV / SubZero-GV & \{1e-3, 5e-3, 1e-2\} & 1e-1 & 4 & \{8, 16\} \\ \hline
\end{tabular}
\end{table}

\begin{table}[t]
\centering
\scriptsize
\caption{Prompts used in our experiments. \textcolor{red}{Red} denotes label words.}
\label{datasets}
\setlength{\tabcolsep}{2pt}
\begin{tabular}{lp{5.5cm}}
\toprule
\textbf{Dataset} & \textbf{Prompt Template} \\ \midrule
SST-2 & \texttt{<text>} It was \textcolor{red}{terrible}/\textcolor{red}{great} \\
RTE & \texttt{<premise>} Does this mean that ``\texttt{<hyp>}'' is true? \textcolor{red}{Yes}/\textcolor{red}{No} \\
CB & Suppose \texttt{<premise>} Can we infer ``\texttt{<hyp>}''? \textcolor{red}{Yes}/\textcolor{red}{No}/\textcolor{red}{Maybe} \\
BoolQ & \texttt{<passage>} \texttt{<question>}? \textcolor{red}{Yes}/\textcolor{red}{No} \\
WSC & \texttt{<text>} Does ``\texttt{<span2>}'' refer to ``\texttt{<span1>}''? \textcolor{red}{Yes}/\textcolor{red}{No} \\
WIC & Does ``\texttt{<word>}'' have the same meaning? \texttt{<s1>} \texttt{<s2>} \textcolor{red}{Yes}/\textcolor{red}{No} \\
MultiRC & \texttt{<para>} Q: \texttt{<q>} Is ``\texttt{<ans>}'' correct? \textcolor{red}{Yes}/\textcolor{red}{No} \\
COPA & \texttt{<premise>} so/because \texttt{<candidate>} \\
ReCoRD & \texttt{<passage>} \texttt{<query>}.replace(``@placeholder'', \texttt{<cand>}) \\
SQuAD & Title: \texttt{<title>} Context: \texttt{<ctx>} Q: \texttt{<q>} Answer: \\
DROP & Passage: \texttt{<ctx>} Q: \texttt{<q>} Answer: \\
\bottomrule
\end{tabular}
\end{table}

\section{Experiments and Analysis}
\noindent{\textbf{Tasks, Models, and Datasets.}} We evaluate on the SuperGLUE \cite{ref31} collection (CB \cite{ref32}, COPA \cite{ref33}, MultiRC \cite{ref34}, RTE \cite{ref35}, WiC \cite{ref36}, WSC \cite{ref37}, BoolQ \cite{ref38}, ReCoRD \cite{ref39}), SST-2 \cite{ref40}, and two QA datasets: SQuAD \cite{ref41} and DROP \cite{ref42}. We test OPT \cite{ref7} (1.3B, 13B, 30B) and Llama2 \cite{ref43} (7B, 13B). As listed in Table \ref{datasets}, the datasets cover classification, multiple-choice, and QA tasks. We consider three fine-tuning schemes: full-tuning (FT), LoRA \cite{ref4}, and prefix-tuning \cite{ref6}.

\noindent{\textbf{Setup.}} We compare with zero-shot, ICL, and Adam fine-tuning, and validate our methods on MeZO \cite{ref8} and SubZero \cite{ref30}. Following MeZO, we sample 1,000/500/1,000 examples for train/val/test. We set $M=4$ and $\alpha=0.5$. MeZO runs for 20K iterations with 2 forward passes per step, while our methods run for 10K iterations with 4 forward passes per step, so that all methods use the same total budget of 40K forward evaluations.
We also include n-SPSA \cite{ref13} as a compute-equivalent baseline. It draws $n$ independent perturbations per step and averages their SPSA estimates, requiring $2n$ forward passes per step. With $n=M/2$, n-SPSA uses the same per-step budget as our methods but allocates it to variance reduction through averaging, whereas our methods use it for directional selection through loss feedback. All models are validated every 1K steps using FP16 on Nvidia A100 80GB or 3090 24GB GPUs. Hyperparameters are listed in Table \ref{tab:unified_params}. 

\noindent{\textbf{Note on Evaluation Protocol.}} Following MeZO \cite{ref8}, we randomly sample subsets from the original datasets for consistent comparison with prior ZO work.

\begin{table*}[t]
\caption{Average task performance on OPT-1.3B (3 rounds). Best per task in \textbf{bold}.}
\label{opt-1.3b_task_performance}
\centering
\setlength{\tabcolsep}{4pt} 
\small
\begin{tabular}{lccccccc|cc|cc}
\toprule
\textbf{Task Type} & \multicolumn{7}{c}{----- classification -----} & \multicolumn{2}{c}{----- multiple choice -----} & \multicolumn{2}{c}{----- generation -----} \\ 
\textbf{Task} & SST2 & RTE & CB & BoolQ & WSC & WIC & MultiRC & COPA & ReCoRD & SQuAD & DROP  \\ \midrule
\textbf{Zero-shot}            & 53.6  & 53.1 & 39.3 & 44.9  & 43.3  & 53.5 & 45.4   & 73.0  & 70.5   & 27.2  & 11.2  \\ 
\textbf{ICL}                  & 80.0  & 53.4 & 44.6 & 59.4  & 46.2  & 50.3  & 46.3   & 69.0  & 71.0   & 58.7  & 20.5  \\ \midrule
\textbf{MeZO(FT)}                 & 89.2  & 57.4 & \textbf{71.4} & 62.5  & 56.7  & 57.2  & 53.3   & 73.0  & 70.9   & 72.0  & 21.9  \\
\textbf{n-SPSA(FT, $n$=2)}     & 89.8  & 58.1 & 70.8 & 63.0  & 57.3  & 57.5  & 54.1   & 73.0  & 71.2   & 72.8  & 22.3  \\
\textbf{MeZO-GV(FT)}               & 93.0  & 60.6 & 69.6 & 64.4  & 60.6  & 58.0  & 58.6   & \textbf{78.0}  & 72.0   & 75.2  & 24.1  \\ \midrule
\textbf{MeZO(LoRA)}            & 90.8  & 61.7 & 71.2 & 63.4  & 58.7  & 60.2  & 57.0   & 74.0  & 71.5   & 77.5  & 23.1  \\ 
\textbf{n-SPSA(LoRA, $n$=2)}   & 91.2  & 62.0 & 70.5 & 63.8  & 59.2  & 60.0  & 57.6   & 74.0  & 71.8   & 77.8  & 23.4  \\
\textbf{MeZO-GV(LoRA)}          & \textbf{93.5}  & 62.8 & 70.5 & \textbf{64.8}  & \textbf{62.5}  & \textbf{60.7}  & 60.6   & 76.0  & 72.4   & 78.7  & 24.4  \\ \midrule
\textbf{MeZO(Prefix)}          & 90.1  & 65.7 & 69.6 & 63.0  & 60.6  & 56.0  & 59.1   & 71.0  & 70.4   & 76.0  & 23.2  \\
\textbf{n-SPSA(Prefix, $n$=2)} & 90.5  & 65.9 & 69.8 & 63.4  & 60.2  & 56.5  & 59.5   & 72.0  & 70.8   & 76.5  & 23.5  \\
\textbf{MeZO-GV(Prefix)}        & 92.1  & \textbf{66.8} & 70.9 & 64.5  & 60.8  & 58.2  & \textbf{62.7}   & 74.0  & \textbf{72.7}   & \textbf{78.8}  & \textbf{24.8}  \\ \bottomrule
\end{tabular}
\end{table*}

\begin{table*}[t]
\caption{Average task performance on OPT-13B (3 rounds). Best ZO performance in \textbf{bold}.}
\label{13b_task_performance}
\centering
\setlength{\tabcolsep}{3pt} 
\small
\begin{tabular}{lccccccc|cc|cc}
\toprule
\textbf{Task Type} & \multicolumn{7}{c}{----- classification -----} & \multicolumn{2}{c}{----- multiple choice -----} & \multicolumn{2}{c}{----- generation -----} \\ 
\textbf{Task} & SST2 & RTE & CB & BoolQ & WSC & WIC & MultiRC & COPA & ReCoRD & SQuAD & DROP  \\ \midrule
\textbf{Zero-shot}            & 58.8  & 59.6 & 46.4 & 59.0  & 38.5  & 55.0  & 46.9   & 80.0  & 81.2   & 46.2  & 14.6  \\ 
\textbf{ICL}                  & 87.0  & 62.1 & 57.1 & 66.9  & 39.4  & 50.5  & 53.1   & 87.0  & 82.5   & 75.9  & 29.6  \\ \midrule
\textbf{ZO-AdaMU (2×)}        & 92.1  & 72.9 & 67.9 & 73.0  & 61.5  & 60.7  & 63.0   & 89.0  & 83.0   & 82.4  & 32.0  \\ 
\textbf{ZO-AdaMU (LoRA)}      & 88.0  & 72.0 & 71.6 & 72.6  & 60.1  & 56.4  & 58.9   & 88.0  & 83.2   & 76.8  & 32.4  \\ 
\textbf{ZO-AdaMU (Prefix)}    & 88.0  & 61.8 & 72.3 & 74.9  & 56.5  & 58.2  & 61.9   & 86.0  & 82.8   & 85.2  & 30.4  \\ \midrule
\textbf{HiZOO}                & 92.1  & 69.3 & 69.4 & 67.3  & 63.5  & 59.4  & 61.3   & 88.0  & 81.4   & 81.9  & 25.0  \\ 
\textbf{HiZOO(LoRA)}          & 90.6 & 67.5 & 69.6 & 70.5  & 63.5  & 60.2  & 60.2   & 87.0  & 81.9   & 83.8  & 25.1  \\ 
\textbf{HiZOO(Prefix)}        & 92.0  & 71.8 & 69.6 & 73.9  & 60.6  & 60.0  & 64.8   & 87.0  & 81.2   & 83.2  & 25.3  \\ \midrule
\textbf{MeZO(FT)}                 & 91.4  & 66.1 & 67.9 & 67.6  & 63.5  & 61.1  & 60.1   & 88.0  & 81.7   & 84.7  & 30.9  \\
\textbf{n-SPSA(FT, $n$=2)}     & 91.9  & 66.8 & 68.5 & 68.2  & 63.8  & 61.4  & 60.5   & 88.0  & 81.9   & 84.5  & 31.2  \\
\textbf{SubZero(FT)}          & 92.1  & 74.0 & 73.2 & 75.3  & 65.4  & 60.8  & 61.0   & 88.0  & 82.3   & 84.5  & 32.0  \\
\textbf{MeZO-GV(FT)}               & 93.9  & 73.5 & 71.6 & 72.5  & 65.4  & 61.4  & 62.5   & 89.0  & 82.9   & 84.9  & 31.7  \\ 
\textbf{SubZero-GV(FT)}          & \textbf{94.7}  & 74.8 & 73.9 & 76.8  & 64.4  & 62.7  & 63.2   & 89.0  & 83.1   & 84.9  & 31.3  \\  \midrule
\textbf{MeZO(LoRA)}            & 89.6  & 67.9 & 66.1 & 73.8  & 64.4  & 59.7  & 61.5   & 84.0  & 81.2   & 83.8  & 31.4  \\ 
\textbf{n-SPSA(LoRA, $n$=2)}   & 90.1  & 68.3 & 66.8 & 74.2  & 64.6  & 59.9  & 61.2   & 85.0  & 81.5   & 83.5  & 31.1  \\
\textbf{SubZero(LoRA)}        & 93.8  & 75.5 & 71.4 & 76.1  & 65.4  & 60.3  & 60.3   & 89.0  & 81.9   & 83.7  & 31.3 \\ 
\textbf{MeZO-GV(LoRA)}          & 91.6  & 72.6 & 72.8 & 75.6  & \textbf{66.3}  & 60.9  & 61.9   & 89.0  & 82.9   & 84.9  & 32.7  \\
\textbf{SubZero-GV(LoRA)}          & 94.0  & 75.8 & 73.8 & \textbf{77.6}  & 65.4  & 63.9  &  64.1  & \textbf{90.0}  & \textbf{83.8}  & \textbf{85.3} &  32.4\\ \midrule
\textbf{MeZO(Prefix)}          & 90.7  & 70.8 & 69.6 & 73.1  & 60.6  & 59.9  & 63.7   & 87.0  & 81.4   & 84.2  & 28.9  \\ 
\textbf{n-SPSA(Prefix, $n$=2)} & 91.0  & 71.2 & 70.1 & 73.5  & 61.0  & 60.2  & 63.4   & 87.0  & 81.7   & 84.0  & 29.3  \\
\textbf{SubZero(Prefix)}          & 91.7  & 73.6 & 80.3 &  76.3 &  62.1 & 61.1  &  63.5  &  88.0 &  82.0 & 83.7 & 32.0 \\
\textbf{MeZO-GV(Prefix)}        & 92.4  & 74.8 & 73.2 & 76.6  & 63.5  & 61.8  & 64.4   & \textbf{90.0}  & 82.7   & 84.3  & 30.9  \\ 
\textbf{SubZero-GV(Prefix)}        & 93.1   &\textbf{76.2}  & \textbf{85.7}  & 77.1 & 64.4 & \textbf{64.1}  & \textbf{65.1}  & 89.0   & 82.5  &  85.1  & \textbf{32.9}    \\ \midrule
\textbf{FT}                   & 92.0  & 70.8 & 83.9 & 77.1  & 63.5  & 70.1  & 71.1   & 79.0  & 74.1   & 84.9  & 31.3  \\ \bottomrule
\end{tabular}
\end{table*}

\begin{table}[t]
\caption{Task Performance Comparison for Different Methods on Llama2-7B.}
\label{tab_llama2_7b}
\centering
\setlength{\tabcolsep}{3pt}
\small
\begin{tabular}{lccccc}
\toprule 
\textbf{Task} & SST2 & RTE & BoolQ & WSC & WIC  \\ \midrule
\textbf{MeZO-10k}  &  85.3 & 58.1 & 72.1 & 60.8  & 57.8  \\
\textbf{MeZO-20k}  & 88.7  & 62.1  & 80.1 &  62.1 &  60.8 \\
\textbf{n-SPSA-10k ($n$=2)} & 86.1 & 58.8 & 73.0 & 61.2 & 58.3 \\
\textbf{MeZO-GV-10k} & 90.4 & 64.3 & 81.3 & 62.5  & 62.3   \\ \midrule
\textbf{MeZO-10k(LoRA)} & 87.7  & 60.6 & 76.9  & 58.9 & 56.3  \\ 
\textbf{MeZO-20k(LoRA)} & 93.7  & 63.3 & 79.5 & 62.5 &  57.5 \\ 
\textbf{n-SPSA-10k(LoRA, $n$=2)} & 88.3 & 61.2 & 77.4 & 59.5 & 57.0 \\
\textbf{MeZO-GV-10k(LoRA)}  & 94.3  & 65.7 & 80.7 & 61.5 & 61.4   \\ \bottomrule
\end{tabular}
\end{table}

\subsection{Medium-sized Language Models}
As shown in Table \ref{opt-1.3b_task_performance}, MeZO-GV consistently outperforms both vanilla MeZO and n-SPSA across all tuning schemes under the same total forward-pass budget. Notable improvements include SST-2 with a 3.8\% gain, COPA with a 5.0\% gain, MultiRC with a 5.3\% gain, and WSC with a 3.9\% gain, where standard methods show limited improvement. MeZO-GV also converges faster: on SST-2 and WSC, it reaches vanilla MeZO's 20K-step accuracy in only 6K and 1K steps, respectively.

\begin{figure*}[t]
\centering
\begin{minipage}{0.28\textwidth} 
\includegraphics[width=\linewidth]{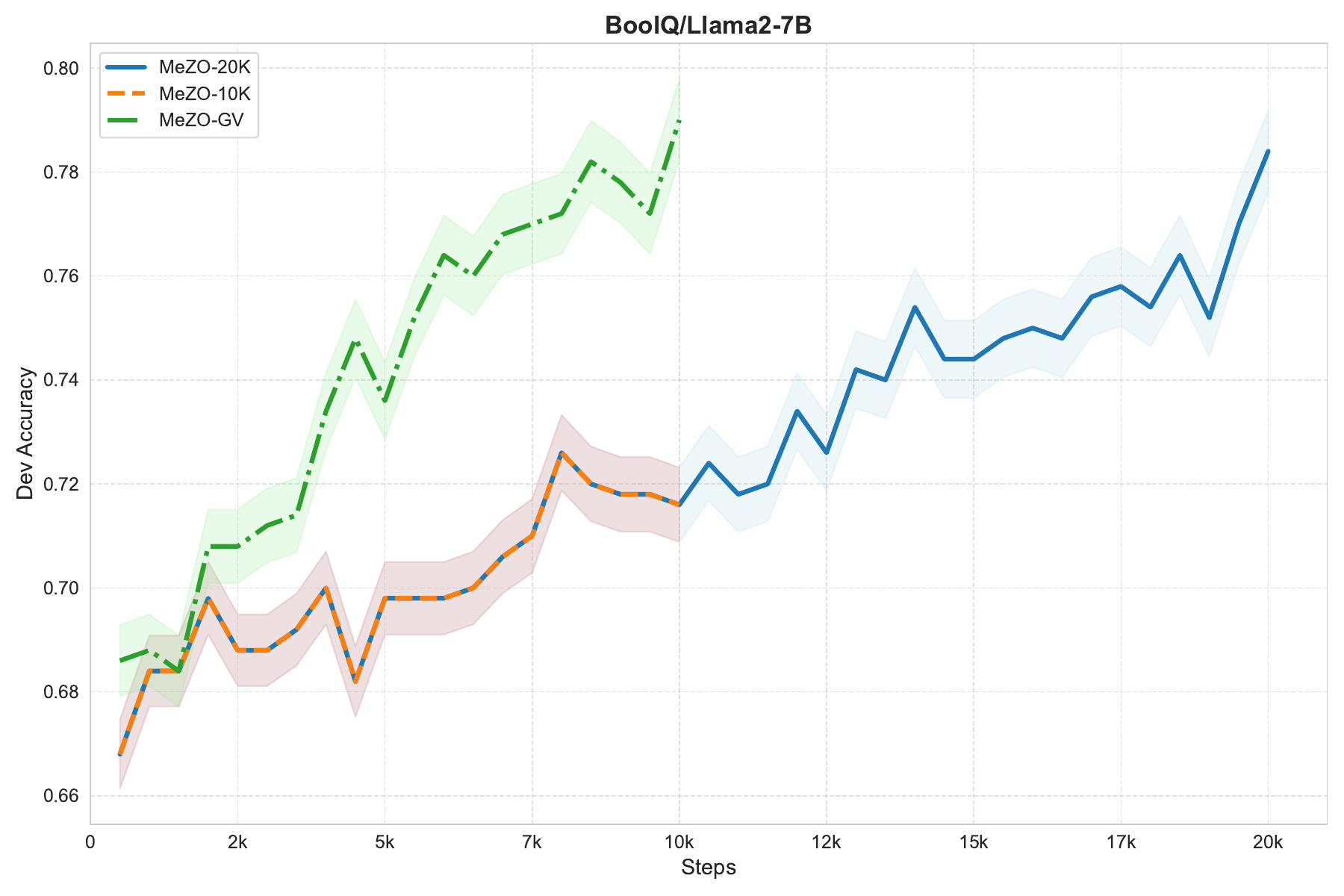}
\label{boolq_llama2_7b_val_acc}
\end{minipage}
\hfill
\begin{minipage}{0.28\textwidth}
\includegraphics[width=\linewidth]{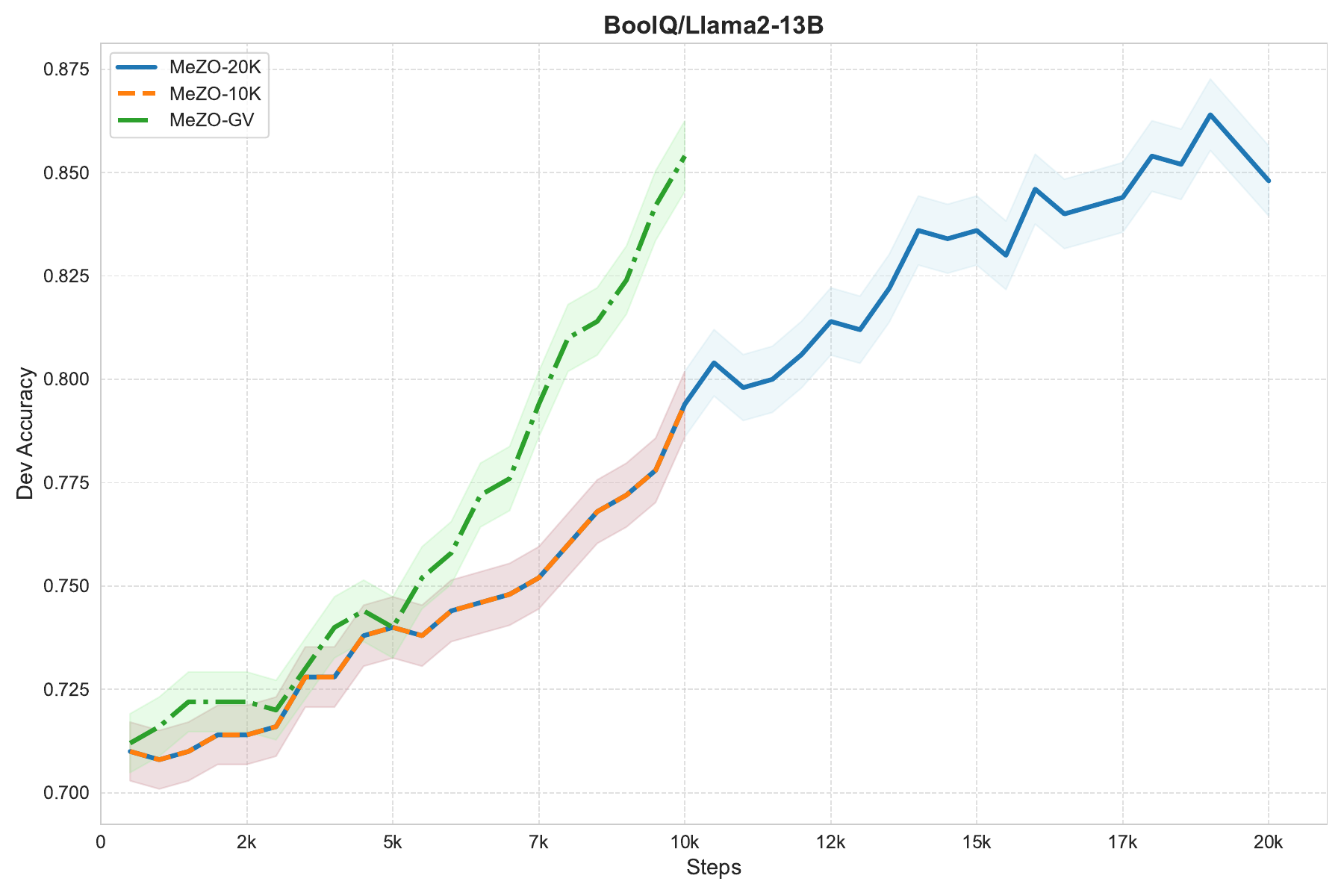}
\label{boolq_llama2_13_val_acc}
\end{minipage}
\hfill
\begin{minipage}{0.28\textwidth}
\includegraphics[width=\linewidth]{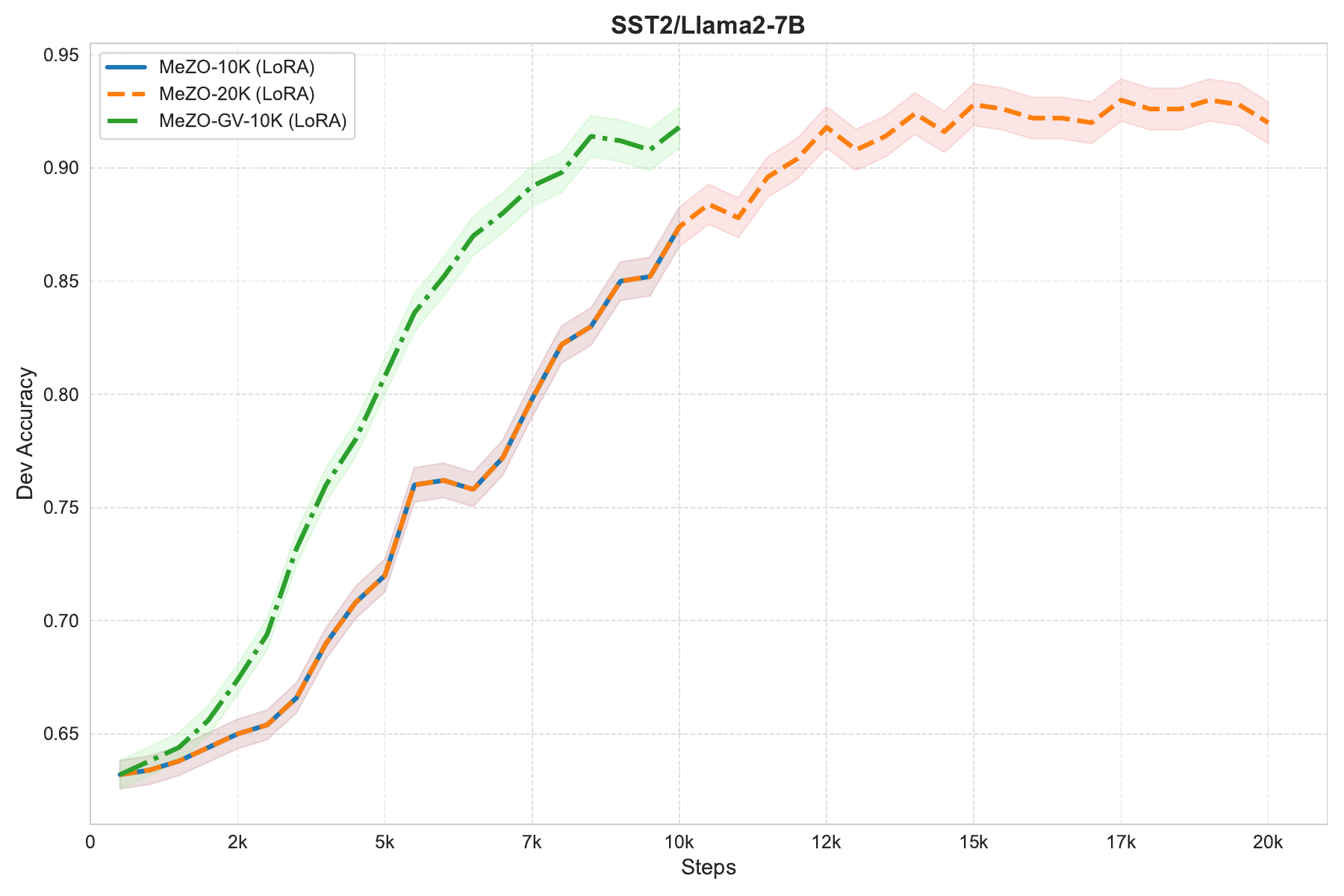}
\label{SST2_val_acc_llama2_7b}
\end{minipage}
\caption{Validation accuracy on SST-2 and BoolQ for Llama2-7B and Llama2-13B. Batch size is 16. The learning rate is 1e-4 for LoRA-based methods and 5e-7 for full-parameter methods.}
\label{llama2_val_acc}
\end{figure*} 
\subsection{Large Language Models}
We scale to larger models to further evaluate the proposed methods. As shown in Table~\ref{13b_task_performance}, GV-based methods consistently outperform their non-GV counterparts and the compute-equivalent n-SPSA ($n$=2) baseline across all tuning schemes on OPT-13B. The consistent advantage over n-SPSA indicates that actively constructing an informed guiding direction is more effective than simply averaging additional random estimates. Among ZO methods, GV-based approaches achieve the best performance on all 11 tasks and outperform gradient-based fine-tuning on 9 of 11 tasks. For example, SubZero-GV with full tuning reaches 94.7\% on SST-2, which is 2.7\% higher than MeZO. SubZero-GV with prefix tuning reaches 85.7\% on CB, improving over ZO-AdaMU by 13.4\%, and 76.2\% on RTE, improving over MeZO by 5.4\%.

Table~\ref{tab_llama2_7b} presents results on Llama2-7B. GV-based methods with 10K steps achieve higher accuracy than MeZO at 20K steps under the same total forward-pass budget of 40K. For instance, MeZO-GV-10k with LoRA achieves 94.3\% on SST-2, outperforming MeZO-20k with LoRA at 93.7\% while using the same computational cost.

Tables~\ref{tab_llama2_13} and \ref{tab_30b_opt} further confirm scalability on Llama2-13B and OPT-30B. On Llama2-13B, MeZO-GV-10k with LoRA achieves 93.7\% on SST-2 and 83.3\% on BoolQ, clearly outperforming both n-SPSA at 90.5\% and 77.1\%, and MeZO at 89.7\% and 76.3\%. On OPT-30B, MeZO-GV with prefix tuning achieves 91.4\% on SST-2 and 77.4\% on BoolQ, compared with n-SPSA at 88.2\% and 74.2\%, and MeZO at 87.5\% and 73.5\%. Across all model scales, directional selection through guiding vectors consistently outperforms variance reduction through averaging.
\begin{table}
\caption{Task Performance Comparison for Different Methods on Llama2-13B}
\label{tab_llama2_13}
\centering
\setlength{\tabcolsep}{4pt}
\small
\begin{tabular}{lccccc}
\toprule 
\textbf{Task} & SST2 & RTE & BoolQ & WSC & WIC  \\ \midrule
\textbf{MeZO-10k(LoRA)} & 89.7  & 66.8 & 76.3  & 59.6 & 59.9  \\ 
\textbf{MeZO-20k(LoRA)} & 94.3  & 70.4 & 82.1 & 61.5 &  62.7 \\ 
\textbf{n-SPSA-10k(LoRA, $n$=2)} & 90.5  & 67.5 & 77.1 & 60.3 & 60.5  \\
\textbf{MeZO-GV-10k(LoRA)}  & 93.7  & 72.2 & 83.3 & 65.4 & 65.8   \\ \bottomrule
\end{tabular}
\end{table}

\begin{table}
\caption{Task Performance Comparison on OPT-30B}
\label{tab_30b_opt}
\centering
\setlength{\tabcolsep}{4pt}
\small
\begin{tabular}{lccccc}
\toprule 
\textbf{Task} & SST2 & RTE & BoolQ & WSC & WIC  \\ \midrule
\textbf{Zero-shot}  & 56.7  & 52.0 & 39.1 & 38.5  & 50.2 \\
\textbf{ICL}  & 81.9  & 66.8  & 66.2 & 56.7  & 51.3  \\ \hline
\textbf{MeZO (prefix)} & 87.5 & 72.6 & 73.5 & 55.7  &59.1    \\ 
\textbf{n-SPSA(prefix, $n$=2)} & 88.2  & 73.1 & 74.2 & 56.3  & 59.8  \\
\textbf{MeZO-GV(prefix)} & 91.4  & 75.8 & 77.4  &  61.5& 62.7  \\  \hline
\textbf{SubZero (prefix)} & 89.3 & 74.0 & 76.8 & 59.6  &58.3    \\ 
\textbf{SubZero-GV(prefix)} & 91.6  & 75.1 & 79.4  &  61.5& 62.9 \\ \bottomrule
\end{tabular}
\end{table}
In Figure~\ref{llama2_val_acc}, the validation accuracy curves further illustrate that GV-based methods reach comparable or higher accuracy with fewer training steps.

\begin{table}[t]
\caption{Task Performance Comparison of Greedy Strategy for Different Methods on Llama2-7B and OPT-13B}
\label{greedy_table_acc}
\centering
\setlength{\tabcolsep}{3pt}
\small
\begin{tabular}{llccc}
\toprule 
\textbf{Model} & \textbf{Task} & \textbf{WiC} & \textbf{RTE} & \textbf{BoolQ} \\ \midrule
\multirow{5}{*}{Llama2-7B} & MeZO          & 60.8 & 62.1 & 80.1 \\
                            & n-SPSA ($n$=2) & 61.2 & 62.5 & 80.5 \\
                            & MeZO-Greedy   & 63.0 & 63.6 & 81.9 \\
                            & MeZO (LoRA)   & 57.5 & 63.3 & 79.5 \\
                            & n-SPSA (LoRA, $n$=2) & 58.1 & 63.8 & 79.9 \\
                            & MeZO-Greedy (LoRA) & 61.9 & 65.7 & 80.8 \\ \midrule
\multirow{5}{*}{OPT-13B}    & MeZO          & 61.1 & 66.1 & 67.6 \\
                            & n-SPSA ($n$=2) & 61.4 & 66.8 & 68.2 \\
                            & MeZO-Greedy   & 61.9 & 72.2 & 72.6 \\
                            & MeZO (LoRA)   & 60.8 & 74.0 & 75.3 \\
                            & n-SPSA (LoRA, $n$=2) & 61.2 & 74.5 & 75.7 \\
                            & MeZO-Greedy (LoRA) & 62.7 & 75.8 & 75.9 \\ \bottomrule
\end{tabular}
\end{table}

\subsection{MeZO with Greedy Strategy}
Table \ref{greedy_table_acc} presents the test accuracy of MeZO, n-SPSA, and MeZO-Greedy on Llama2-7B and OPT-13B. The Greedy variants consistently outperform both n-SPSA and standard MeZO under the same forward-pass budget. On OPT-13B, MeZO-Greedy improves RTE from 66.1\% to 72.2\%, compared with 66.8\% for n-SPSA, and improves BoolQ from 67.6\% to 72.6\%, compared with 68.2\% for n-SPSA. When combined with LoRA, the Greedy variants maintain or further improve accuracy, confirming the robustness of the greedy selection strategy.

\subsection{Impact of the Number of Evaluations}
\begin{figure}[ht]
\centering
\begin{minipage}{0.3\textwidth} 
\includegraphics[width=\linewidth]{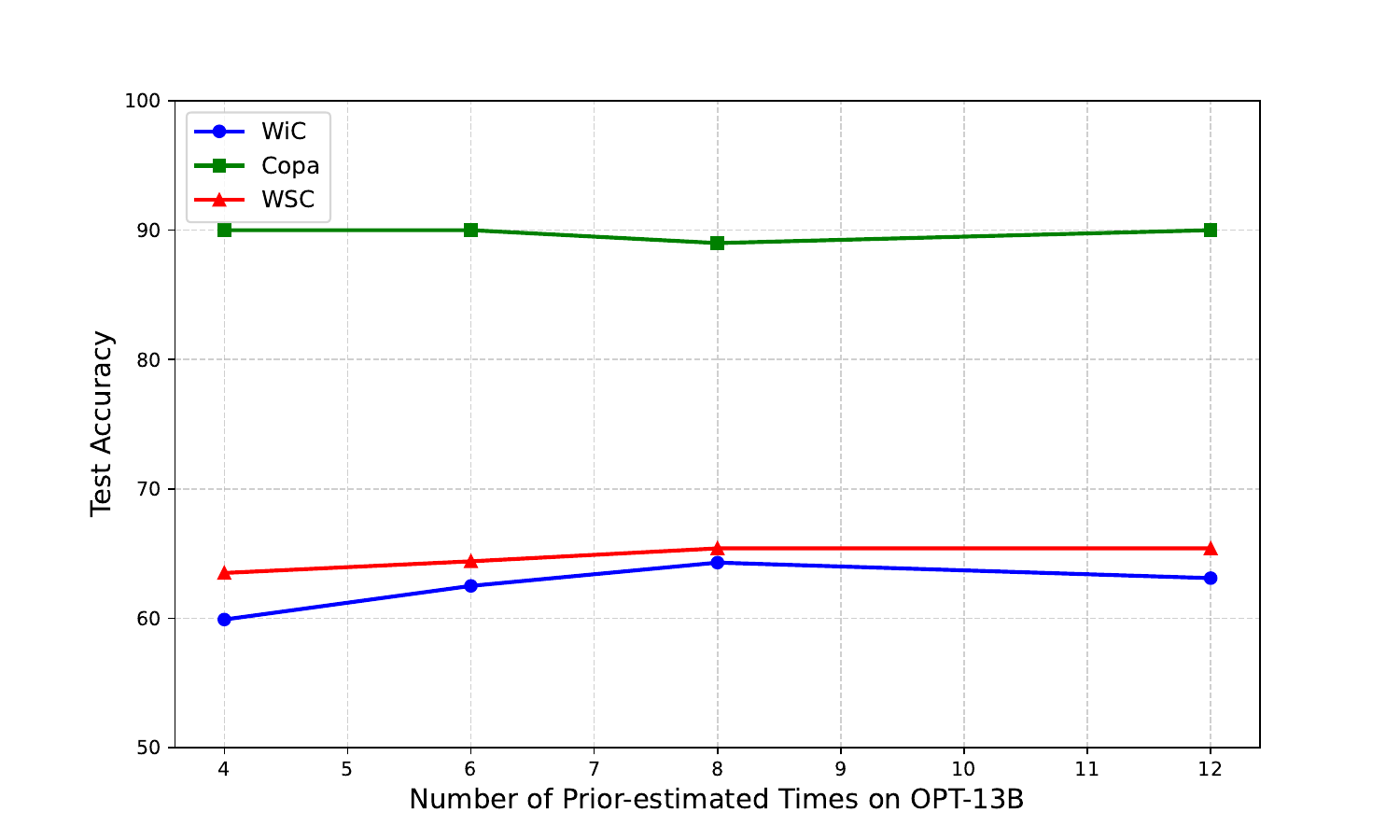}
\end{minipage}
\caption{OPT-13B performance on three datasets as a function of the number of candidate evaluations.}
\label{fwa_times}
\end{figure}

Figure \ref{fwa_times} shows OPT-13B performance as the number of candidate evaluations varies from 4 to 12. COPA and WSC exhibit relatively stable performance, while WIC shows the most notable improvement, suggesting that for most tasks, a small number of candidate evaluations is sufficient.
\subsection{Memory Usage of Different Methods}
Table \ref{tab:fine_tuning_memory} compares memory usage for fine-tuning OPT-13B. MeZO-GV variants use the same amount of memory as MeZO, ranging from 28 to 47\,GB, while improving convergence. In contrast, full fine-tuning requires 242 to 315\,GB.
\begin{table}[htbp]
\centering
\small
\caption{Memory usage (GB) of fine-tuning OPT-13B, with FT\textquotesingle s batch size being 8 and 16 for other tasks.}
\label{tab:fine_tuning_memory}
\begin{tabular}{lcccc}
\toprule
\multirow{2}{*}{Method} & \multicolumn{3}{c}{Task} \\
\cmidrule(lr){2-4}
 & SST-2 & WIC & BoolQ \\
\midrule
Zero-shot       & 26.0 & 26.0 & 26.3 \\
ICL             & 27.2 & 28.5 & 29.3 \\
FT      & 242.3 & 244.7 & 315.3 \\
MeZO (FT)       & 28.9 & 29.1 & 45.6 \\
MeZO (LoRA)     & 28.6 & 29.3 & 46.5 \\
MeZO (Prefix)   & 29.5 & 29.7 & 46.9 \\
MeZO-GV (FT)    & 28.9 & 29.1 & 45.6 \\
MeZO-GV (LoRA)  & 28.6 & 29.3 & 46.5 \\
MeZO-GV (Prefix)& 29.5 & 29.7 & 46.9 \\
\bottomrule
\end{tabular}
\end{table}

\subsection{Directional Alignment Analysis}\label{cos_sim}
\begin{figure}[t]
\centering
\begin{minipage}{0.22\textwidth} 
\includegraphics[width=\linewidth]{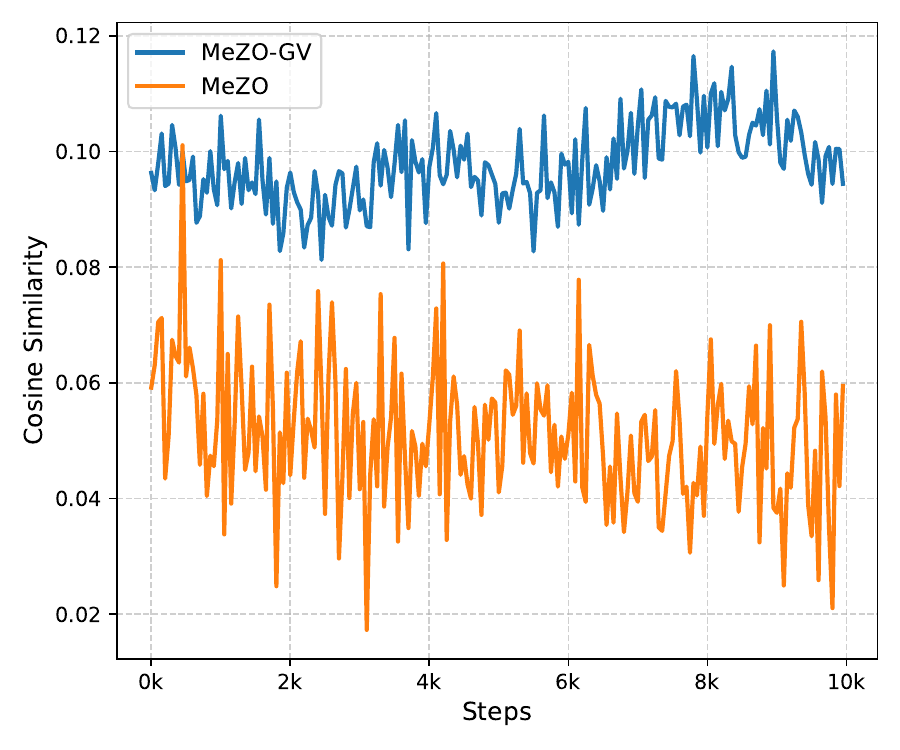}
\label{SST2_cos_sim_1.3b}
\end{minipage}
\hfill
\begin{minipage}{0.22\textwidth}
\includegraphics[width=\linewidth]{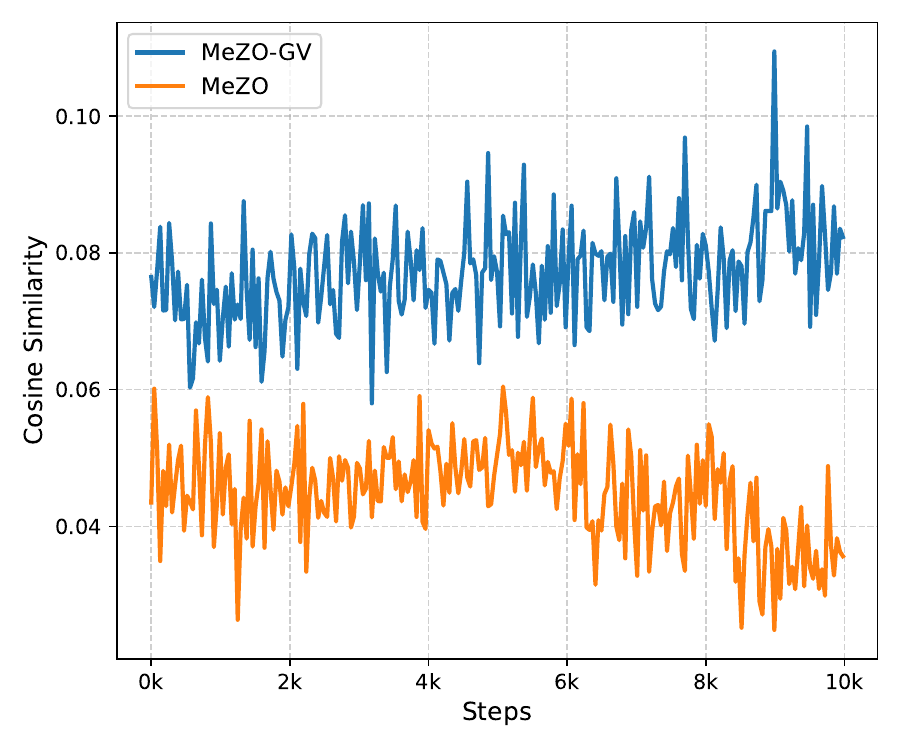}
\label{boolq_cos_sim_1.3b}
\end{minipage}
\caption{Cosine similarity between the estimated gradient \(\hat{\mathbf{g}}\) and the true gradient \(\mathbf{g}\) computed by SGD, on SST-2 and BoolQ using OPT-1.3B in the prefix tuning scheme.}
\label{fig_cos_sim}
\end{figure}

We measure the cosine similarity between the estimated gradient $\hat{\mathbf{g}}$ and the true gradient $\mathbf{g}$ where the true gradient is computed via SGD. As shown in Figure \ref{fig_cos_sim}, MeZO-GV consistently achieves higher cosine similarity than standard MeZO on SST-2 and BoolQ using OPT-1.3B with prefix tuning, providing empirical support for our theoretical analysis.

\subsection{Comparison with n-SPSA}
\label{sec:nspsa}

We compare against n-SPSA \cite{ref13}, which generalizes MeZO by drawing $n$ independent perturbations and averaging their SPSA estimates, requiring $2n$ forward passes per step. While n-SPSA reduces variance through averaging, our methods use the same budget for directional selection through loss feedback. Table~\ref{tab:nspsa_comparison} presents results on OPT-1.3B over 10K steps. Our method consistently outperforms n-SPSA at all budget levels. With 6 forward passes per step, MeZO-GV achieves 93.5\% on SST-2 versus 89.8\% for n-SPSA with $n=3$, showing that loss-guided direction construction is more effective than simple averaging.

\begin{table}[t]
\centering
\caption{Performance comparison with n-SPSA under identical computational budgets on OPT-1.3B with 10K steps.}
\label{tab:nspsa_comparison}
\scriptsize
\begin{tabular}{llcccc}
\toprule
\multirow{2}{*}{\textbf{FP/Step}} & \multirow{2}{*}{\textbf{Method}} & Total & SST-2 & WSC & BoolQ \\
\cmidrule(lr){4-4} \cmidrule(lr){5-5} \cmidrule(lr){6-6}
& & FP & Acc & Acc & Acc \\
\midrule
\multirow{2}{*}{4} & n-SPSA ($n=2$) & 40K & 89.6 & 56.7 & 62.5 \\
& \textbf{MeZO-GV} & 40K & \textbf{91.8} & \textbf{58.7} & \textbf{63.2} \\
\midrule
\multirow{2}{*}{6} & n-SPSA ($n=3$) & 60K & 89.8 & 57.1 & 62.6 \\
& \textbf{MeZO-GV} & 60K & \textbf{93.5} & \textbf{60.6} & \textbf{64.4} \\
\midrule
\multirow{2}{*}{8} & n-SPSA ($n=4$) & 80K & 89.9 & 57.7 & 62.8 \\
& \textbf{MeZO-GV} & 80K & \textbf{93.7} & \textbf{62.5} & \textbf{64.7} \\
\bottomrule
\end{tabular}
\end{table}

\subsection{Wall-clock Time Analysis}\label{sec:wallclock}
Table~\ref{tab:wallclock} reports wall-clock training time on SST-2 using OPT-1.3B for 10K steps on a single A100 GPU. Our method reduces total training time by approximately 29\% compared to n-SPSA at 6 forward passes per step, achieving 0.84h versus 1.18h. This improvement comes from the higher cosine similarity between the guiding vector and the true gradient, which leads to larger per-step loss reductions that compensate for the overhead of constructing the guiding vector.

\begin{table}[h]
\centering
\caption{Wall-clock Time Comparison on OPT-1.3B for the SST-2 Task. We report the average time per step (seconds) and total training time (hours) for 10K steps on a single NVIDIA A100 GPU.}
\label{tab:wallclock}
\scriptsize
\begin{tabular}{llccc}
\toprule
\textbf{FP/Step} & \textbf{Method} & \textbf{Time/Step} & \textbf{Total Time} & \textbf{Acc.} \\
\midrule
\multirow{2}{*}{4} & n-SPSA & 0.33s & 0.92h & 89.6 \\
                   & Ours   & \textbf{0.32s} & \textbf{0.88h} & \textbf{91.8} \\
\midrule
\multirow{2}{*}{6} & n-SPSA & 0.42s & 1.18h & 89.8 \\
                   & Ours   & \textbf{0.30s} & \textbf{0.84h} & \textbf{93.5} \\
\midrule
\multirow{2}{*}{8} & n-SPSA & 0.52s & 1.44h & 89.9 \\
                   & Ours   & 0.55s & 1.52h & \textbf{93.7} \\
\bottomrule
\end{tabular}
\end{table}

\subsection{Hyperparameter Sensitivity Analysis}\label{sec:ablation}
We conduct ablation studies to examine the sensitivity of our method to two key hyperparameters: the number of candidate perturbations $M$ and the selection ratio $\alpha$.

\noindent\textbf{Effect of $M$.} Table~\ref{tab:ablation_M} presents MeZO-GV performance with varying $M$ on OPT-13B. The results indicate that $M=4$ provides a good balance between computational overhead and performance.

\begin{table}[h]
\centering
\caption{Effect of $M$ on MeZO-GV Performance (OPT-13B)}
\label{tab:ablation_M}
\small
\begin{tabular}{lcccc}
\toprule
\textbf{Method} & \textbf{$M$ (Query Count)} & \textbf{SST2} & \textbf{WSC} & \textbf{COPA} \\
\midrule
\multirow{5}{*}{MeZO-GV} & 4  & 93.9 & 65.4 & 89 \\
                         & 6  & 93.8 & 63.5 & 88 \\
                         & 8  & 92.4 & 64.4 & 89 \\
                         & 10 & 94.7 & 65.4 & 91 \\
                         & 14 & 94.3 & 64.4 & 90 \\
\bottomrule
\end{tabular}
\end{table}

\noindent\textbf{Effect of $\alpha$.} Table~\ref{tab:ablation_alpha} examines the effect of $\alpha$ on MeZO-GV. Overall, $\alpha = 0.5$ yields the most stable performance. Smaller values of $\alpha$ (e.g., 0.2) increase the variance of the guiding vector, while larger values (e.g., 0.7) dilute the signal from informative directions.

\begin{table}[h]
\centering
\caption{Effect of $\alpha$ on MeZO-GV Performance (OPT-13B, $M$=14)}
\label{tab:ablation_alpha}
\small
\begin{tabular}{lccccc}
\toprule
\textbf{$\alpha$} & \textbf{SST-2} & \textbf{WSC}  & \textbf{COPA}  \\
\midrule
0.2 & 92.1 & 63.5 & 87 \\
0.3 & 94.3  & 64.4 & 90  \\
0.5 & 93.9  & 65.4 & 89  \\
0.7 & 91.4  & 61.5 & 87  \\
\bottomrule
\end{tabular}
\end{table}

\subsection{Detailed Theoretical Derivations}
\label{sec:appendix_proofs}

In this subsection, we present the full derivations for the convergence analysis and the alignment properties of the proposed methods, following the Greedy Descent Framework \cite{ref46}.

\subsubsection{Preliminaries}
\label{sec:appendix_prelim}
We assume $\mathcal{L}$ is $L$-smooth and convex. $L$-smoothness implies that for all $x,y$:
\begin{equation}
\label{eq:appendix_smoothness}
    \mathcal{L}(y) \leq \mathcal{L}(x) + \nabla \mathcal{L}(x)^\top (y-x) + \frac{L}{2}\|y-x\|^2.
\end{equation}
Convexity gives $\mathcal{L}(y) \geq \mathcal{L}(x) + \nabla \mathcal{L}(x)^\top (y-x)$, which together with the Cauchy--Schwarz inequality yields:
\begin{equation}
\label{eq:appendix_convex_grad_bound}
    \mathcal{L}(x) - \mathcal{L}(\theta^*) \leq \|\nabla \mathcal{L}(x)\|\,\|x-\theta^*\|.
\end{equation}
For a unit direction $v$ and step $\epsilon>0$, the finite-difference estimator $g_\epsilon(v;x) := (\mathcal{L}(x+\epsilon v) - \mathcal{L}(x))/\epsilon$ satisfies
\begin{equation}
    |g_\epsilon(v;x) - \nabla \mathcal{L}(x)^\top v| \leq L\epsilon/2
\end{equation}
by applying Eq.~\eqref{eq:appendix_smoothness} in both directions. This $O(\epsilon)$ accuracy justifies the use of directional derivatives in place of finite differences in the analysis below.

\subsubsection{Proof of Convergence (Theorem \ref{thm:convergence_rate})}

We derive the convergence rate of the generic update rule $\theta_{t+1} = \theta_t - \eta (\nabla \mathcal{L}(\theta_t)^\top v_t) v_t$.

\paragraph{Descent Inequality.}
Substituting $\Delta_t = -\eta (\nabla \mathcal{L}(\theta_t)^\top v_t) v_t$ into Eq.~\eqref{eq:appendix_smoothness} with $\|v_t\|=1$:
\begin{align}
    \mathcal{L}(\theta_{t+1}) &\leq \mathcal{L}(\theta_t) + \nabla \mathcal{L}(\theta_t)^\top \Delta_t + \frac{L}{2} \|\Delta_t\|^2 \nonumber \\
    &= \mathcal{L}(\theta_t) - \eta (\nabla \mathcal{L}(\theta_t)^\top v_t)^2 \nonumber \\
    &\quad + \frac{L\eta^2}{2} (\nabla \mathcal{L}(\theta_t)^\top v_t)^2 \nonumber \\
    &= \mathcal{L}(\theta_t) - \Big( \eta - \frac{L \eta^2}{2} \Big) (\nabla \mathcal{L}(\theta_t)^\top v_t)^2.
\end{align}
Setting $\eta^* = 1/L$ maximizes the coefficient $\eta - L\eta^2/2$, yielding $1/(2L)$. Defining $C_t = (\nabla \mathcal{L}(\theta_t)^\top v_t)^2/\|\nabla \mathcal{L}(\theta_t)\|^2$:
\begin{equation}
    \mathcal{L}(\theta_{t+1}) \leq \mathcal{L}(\theta_t) - \frac{C_t}{2L} \|\nabla \mathcal{L}(\theta_t)\|^2.
    \label{eq:descent_lemma}
\end{equation}

\paragraph{Gradient Norm and Suboptimality.}
Let $\delta_t = \mathcal{L}(\theta_t) - \mathcal{L}(\theta^*)$. From Eq.~\eqref{eq:descent_lemma}: $\delta_{t+1} \leq \delta_t - \frac{C_t}{2L} \|\nabla \mathcal{L}(\theta_t)\|^2$.
By convexity and the Cauchy--Schwarz inequality:
\begin{equation}
    \delta_t \leq \nabla \mathcal{L}(\theta_t)^\top (\theta_t - \theta^*) \leq \|\nabla \mathcal{L}(\theta_t)\| \|\theta_t - \theta^*\|.
\end{equation}
Since the descent inequality ensures $\mathcal{L}(\theta_t) \leq \mathcal{L}(\theta_0)$ for all $t$ (by induction), the iterates remain within the initial level set $\mathcal{S}_0 = \{\theta : \mathcal{L}(\theta) \leq \mathcal{L}(\theta_0)\}$. Let $R = \sup_{\theta \in \mathcal{S}_0} \|\theta - \theta^*\|$ denote the diameter of $\mathcal{S}_0$. We then have $\|\theta_t - \theta^*\| \leq R$ for all $t$, and thus:
\begin{equation}
    \|\nabla \mathcal{L}(\theta_t)\|^2 \geq \frac{\delta_t^2}{R^2}.
\end{equation}

\paragraph{Solving the Recurrence.}
Substituting the gradient bound into the descent inequality:
\begin{equation}
    \delta_{t+1} \leq \delta_t - \frac{C_t}{2L R^2} \delta_t^2.
\end{equation}
Taking expectations (noting that $\mathbb{E}[C_t]$ depends only on $d, M, \alpha$ and is independent of $\theta_t$ under isotropic Gaussian sampling) and applying Jensen's inequality $\mathbb{E}[\delta_t^2] \geq (\mathbb{E}[\delta_t])^2$, we obtain:
\begin{equation}
    \mathbb{E}[\delta_{t+1}] \leq \mathbb{E}[\delta_t] - \frac{\mathbb{E}[C_t]}{2L R^2} (\mathbb{E}[\delta_t])^2.
\end{equation}
To solve this recurrence, we use the algebraic inequality: for $x > 0$ and $0 < cx < 1$, $\frac{1}{x - cx^2} \geq \frac{1}{x} + c$. This follows by setting $a=cx$, which reduces to $\frac{1}{1-a} \geq 1+a$, valid since $(1-a)(1+a)=1-a^2 \leq 1$. Applying this with $c = \mathbb{E}[C_t]/(2LR^2)$ (valid when $\mathbb{E}[\delta_t] < 2LR^2/\mathbb{E}[C_t]$, which is naturally satisfied as the gap decreases):
\begin{equation}
    \frac{1}{\mathbb{E}[\delta_{t+1}]} \geq \frac{1}{\mathbb{E}[\delta_t]} + \frac{\mathbb{E}[C_t]}{2L R^2}.
\end{equation}
Telescoping from $t=0$ to $T-1$ and dropping the positive term $1/\mathbb{E}[\delta_0]$:
\begin{equation}
    \mathbb{E}[\delta_T] \leq \frac{2L R^2}{\sum_{t=0}^{T-1} \mathbb{E}[C_t]}.
\end{equation}
This completes the proof. \qed

\subsubsection{Derivation of Alignment Bounds}
\label{subsec:alignment_bounds}

In this subsection, we derive the expected squared cosine similarity $\mathbb{E}[C_t]$ for each method. Let $g = \nabla \mathcal{L}(\theta_t)$ denote the true gradient. Without loss of generality, we align the coordinate system so that $g = \|g\| e_1$. For any perturbation $z = (z_1, z_2, \ldots, z_d)^\top$, its projection onto the gradient direction is then $z_1$.

\paragraph{Standard MeZO (Baseline).}
For $z \sim \mathcal{N}(0, I_d)$ and $v = z/\|z\|$, the squared cosine similarity is:
\begin{equation}
    C_t = \frac{(g^\top z)^2}{\|g\|^2 \|z\|^2} = \frac{z_1^2}{\sum_{j=1}^d z_j^2}.
\end{equation}
Since each $z_j \sim \mathcal{N}(0, 1)$ i.i.d., by exchangeability $\mathbb{E}[z_k^2/\sum_j z_j^2]$ is identical for all $k$, and summing over $k$ gives 1, hence:
\begin{equation}
    \boxed{\mathbb{E}[C_t^{\text{MeZO}}] = \frac{1}{d}.}
\end{equation}
For $d \sim 10^9$, this value is extremely small ($\sim 10^{-9}$), which explains the slow convergence of standard ZO methods.

\paragraph{MeZO-Greedy.}
The greedy strategy samples $M-2$ candidates $z^{(i)} \sim \mathcal{N}(0, I_d)$ and selects $k^* = \arg\min_i \mathcal{L}(\theta + \epsilon z^{(i)})$. Under a first-order Taylor expansion, this is equivalent to selecting the perturbation with the minimum projection $Y_{(1)} = \min_i z_1^{(i)}$ onto the gradient direction.

We decompose $z^* = Y_{(1)} e_1 + z^*_\perp$, giving $C_t = Y_{(1)}^2/(Y_{(1)}^2 + \|z^*_\perp\|^2)$.

\textit{Signal Term.}
From extreme value theory \cite{ref47}, the CDF of $Y_{(1)}$ is $F_{Y_{(1)}}(y) = 1 - [1 - \Phi(y)]^{M-2}$.
Setting $F_{Y_{(1)}}(y_{\text{med}}) = 1/2$ and applying the Mills ratio $1 - \Phi(y) \approx \phi(y)/|y|$ for $|y| \gg 1$:
\begin{align}
    &[1-\Phi(y_{\text{med}})]^{M-2} = \tfrac{1}{2} \nonumber \\
    &\Longrightarrow\;
    \Phi(y_{\text{med}}) \approx \tfrac{\ln 2}{M-2}
    \;\Longrightarrow\;
    y_{\text{med}}^2 \approx 2\log(M-2),
\end{align}
where the second step uses $1-(1-x)^n \approx nx$ for small $x$, and the third applies the Mills ratio.
More precisely, $Y_{(1)}$ converges to a Gumbel distribution:
\begin{equation}
    Y_{(1)} \xrightarrow{d} -\sqrt{2 \log (M\!-\!2)} + \frac{W}{\sqrt{2 \log (M\!-\!2)}},
\end{equation}
where $W \sim \text{Gumbel}$. Thus $\mathbb{E}[Y_{(1)}^2] \approx 2\log(M-2) + O(1/\log(M-2)) \approx 2 \log (M-2)$.

\textit{Noise Term.}
Since the selection depends only on $\{z_1^{(i)}\}$ and $z_1^{(i)} \perp z_\perp^{(i)}$ in $\mathcal{N}(0, I_d)$, the orthogonal part retains $z^*_\perp \sim \mathcal{N}(0, I_{d-1})$ after conditioning, so $\mathbb{E}[\|z^*_\perp\|^2] = d - 1$.

\textit{Expected Alignment.}
Since $\|z^*_\perp\|^2 \sim \chi^2_{d-1}$ concentrates around $d-1$ for $d \gg 1$, the ratio approximation (with $O(1/d)$ error by the delta method) yields:
\begin{equation}
    \mathbb{E}[C_t^{\text{Greedy}}] \approx \frac{2 \log (M\!-\!2)}{2 \log (M\!-\!2) + (d-1)}.
\end{equation}
For $d \gg \log M$:
\begin{equation}
    \boxed{\mathbb{E}[C_t^{\text{Greedy}}] \approx \frac{2 \log (M-2)}{d}.}
\end{equation}
This gives a speedup factor of $2 \log (M-2)$ over standard MeZO. For $M = 10$, which corresponds to 8 candidates, this is approximately $4\times$.

\paragraph{MeZO-GV (Guiding Vector).}
MeZO-GV samples $M-2$ perturbations and constructs a guiding vector from the top and bottom performers. Let $K = \lfloor \alpha (M-2) \rfloor$ with $\alpha \in (0, 0.5]$. The guiding vector is:
\begin{equation}
    v_{\text{raw}} = \frac{1}{K} \sum_{i \in S_1} z^{(i)} - \frac{1}{K} \sum_{j \in S_2} z^{(j)},
\end{equation}
where $S_1$ (bottom $K$) and $S_2$ (top $K$) are selected by ranking the projections $y_i = z_1^{(i)}$.

\textit{Signal Component.}
We decompose $v_{\text{raw}} = v_{\text{raw}, \parallel} + v_{\text{raw}, \perp}$. The first component is $(v_{\text{raw}})_1 = \bar{Y}_{\text{bottom}} - \bar{Y}_{\text{top}}$.
The truncated mean of the bottom $\alpha$-quantile of $\mathcal{N}(0,1)$ is:
\begin{equation}
    \mu_\alpha^- = \frac{1}{\alpha} \int_{-\infty}^{\Phi^{-1}(\alpha)} \!\! y \, \phi(y) \, dy = -\frac{\phi(\Phi^{-1}(\alpha))}{\alpha},
\end{equation}
which is strictly negative, where we used $y\phi(y) = -\phi'(y)$. By symmetry $\mu_\alpha^+ = -\mu_\alpha^-$, so:
\begin{equation}
    \mathbb{E}[(v_{\text{raw}})_1] \approx 2\mu_\alpha^- < 0,
\end{equation}
confirming descent alignment.

\textit{Signal Energy.}
By bias-variance decomposition:
\begin{equation}
    \mathbb{E}[(v_{\text{raw}})_1^2] = 4(\mu_\alpha^-)^2 + \frac{2\sigma_\alpha^2}{K},
\end{equation}
where $\sigma_\alpha^2 = \text{Var}(Y \mid Y \leq \Phi^{-1}(\alpha))$. The additive form relies on approximate independence between $\bar{Y}_{\text{bottom}}$ and $\bar{Y}_{\text{top}}$, justified by the $(1-2\alpha)(M-2)$ unselected samples that decouple the two groups. For large $K$, the signal energy is dominated by the squared mean: $\mathbb{E}[\|v_{\text{raw}, \parallel}\|^2] \approx 4(\mu_\alpha^-)^2$.

\textit{Noise Component.}
The orthogonal component $v_{\text{raw}, \perp}$ has $\mathbb{E}[v_{\text{raw}, \perp}] = 0$. Because the selection of $S_1, S_2$ depends only on $\{z_1^{(i)}\}$ and $z_1^{(i)} \perp z_\perp^{(i)}$, each $z_\perp^{(i)}$ retains $\mathcal{N}(0, I_{d-1})$ after conditioning. With $S_1, S_2$ disjoint, $\text{Var}(v_{\text{raw}, \perp}) = \frac{2}{K} I_{d-1}$:
\begin{equation}
    \mathbb{E}[\|v_{\text{raw}, \perp}\|^2] = \text{tr}\Big(\tfrac{2}{K} I_{d-1}\Big) = \frac{2(d\!-\!1)}{K} \approx \frac{2d}{K}.
\end{equation}

\textit{Expected Alignment.}
Since $\|v_{\text{raw}, \perp}\|^2$ concentrates around $2d/K$ for $d \gg 1$, the ratio approximation (with $O(1/d)$ error by the delta method) gives:
\begin{align}
    \mathbb{E}[C_t^{\text{GV}}] &\approx \frac{4(\mu_\alpha^-)^2}{4(\mu_\alpha^-)^2 + 2d/K} \nonumber \\
    &\approx \frac{2K(\mu_\alpha^-)^2}{d} \quad (d \gg K(\mu_\alpha^-)^2).
\end{align}
Substituting $K = \lfloor\alpha (M-2)\rfloor$:
\begin{equation}
    \boxed{\mathbb{E}[C_t^{\text{GV}}] \approx \frac{2 K (\mu_\alpha^-)^2}{d}.}
\end{equation}
For $M=4$, $\alpha=0.5$, and $K=1$, the guiding vector $v = z_{\text{low}} - z_{\text{high}}$ doubles the gradient signal amplitude, since $\mathbb{E}[v] \propto -2g$ compared to $\mathbb{E}[z] \propto -g$ for a single perturbation. This can be viewed as signal amplification that speeds up training.

\section{Conclusion}
We have presented two complementary approaches that turn random perturbations into more effective descent directions for zeroth-order optimization: a guiding vector strategy and a greedy perturbation strategy. By using loss feedback to refine update directions, both methods provide a larger per-iteration reduction in the objective than standard MeZO, as confirmed by our theoretical analysis. Experiments across LLMs of different scales show that the proposed methods converge faster and achieve higher accuracy. These results suggest that using loss evaluations to guide perturbation selection is a practical way to address the inefficiency of random sampling in high-dimensional spaces, providing a scalable approach to memory-efficient LLM fine-tuning.

\vfill

\end{document}